\documentclass[letterpaper, 10 pt, conference]{ieeeconf}  % Comment this line out if you need a4paper

\IEEEoverridecommandlockouts                              % This command is only needed if 
\usepackage{amsmath,amssymb,amsfonts}
\usepackage{algorithmic}
\usepackage{graphics,graphicx,color}
\usepackage{textcomp}
\usepackage{xcolor}
\usepackage{algorithm}
\usepackage{algorithmic}
\usepackage{graphics} % for pdf, bitmapped graphics files
\usepackage{tabularx}

\usepackage{caption}

\title{\LARGE \bf
Continuous Control of Diverse Skills in Quadruped Robots Without Complete Expert Datasets
}

\author{\small Jiaxin Tu$^{1, \dagger}$, Xiaoyi Wei$^{1, \dagger}$, Yueqi Zhang$^{1}$, Taixian Hou$^{1}$, Xiaofei Gao$^{2}$, Zhiyan Dong$^{1}$, Peng Zhai$^{1,*}$ and Lihua Zhang$^{3,*}$% <-this % stops a space
\thanks{$^{1}$Jiaxin Tu, Xiaoyi Wei, Yueqi Zhang, Taixian Hou, Zhiyan Dong, and Peng Zhai are with the Academy for Engineering and Technology, Fudan University, Shanghai 200433, China
        {\tt\small \{jxtu22, weixy23, zhangyq23, txhou21\}@m.fudan.edu.cn; \{dongzhiyan, pzhai\}@fudan.edu.cn}}%
\thanks{$^{2}$Xiaofei Gao is with Beijing Jingcheng Zhitong Robotics Technology Co., Beijing, China
        {\tt\small gaoxiaofei@inter-smart.com}}%
\thanks{$^{3}$Lihua Zhang is with the Engineering Research Center of AI and Robotics, Shanghai, China
        {\tt\small lihuazhang@fudan.edu.cn}}%
\thanks{$\dagger$ Equal Contribution.}
\thanks{* Corresponding Author.}
}

\let\oldtwocolumn\twocolumn
\renewcommand\twocolumn[1][]{%
\oldtwocolumn[{#1}{
\begin{center}
\vspace{-0.6cm}
\captionsetup{font=small}
\includegraphics[width=\textwidth]{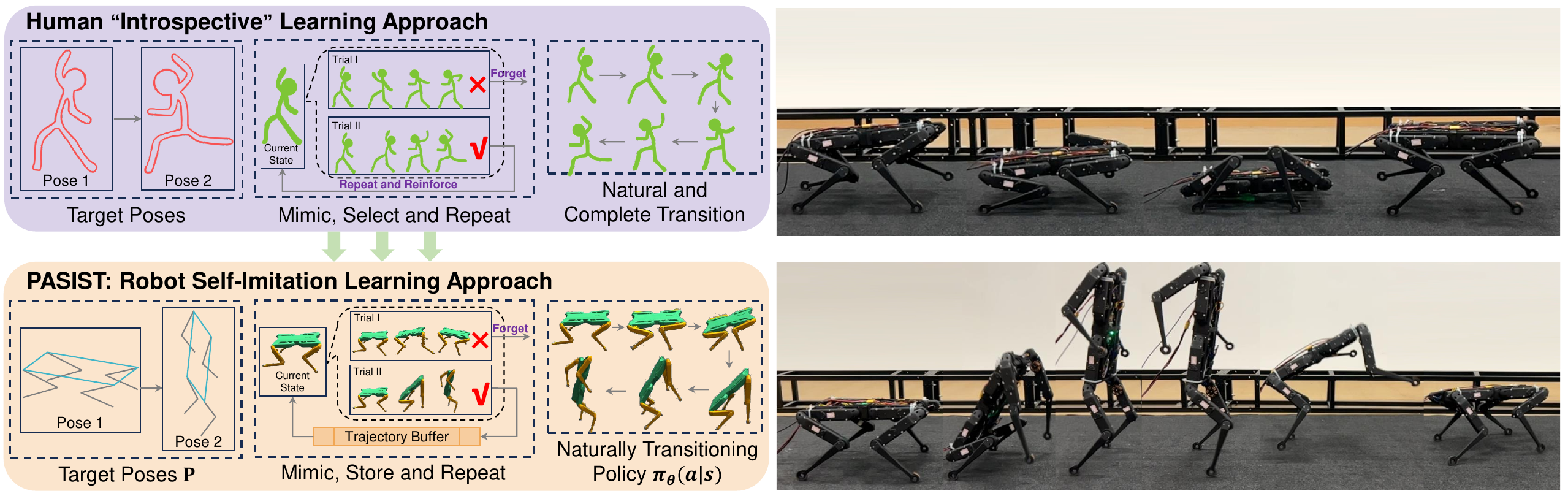}

\captionof{figure}{\textbf{Left}: Inspiration of PASIST. Humans learn new skills by repeatedly imitating, exploring, discarding incorrect trajectories, and reinforcing correct ones, even with only a target pose available. Following this approach, PASIST uses the target pose as the learning objective and stores high-quality trajectories discovered during exploration, enabling skill acquisition and smooth transitions. \textbf{Right}: Sim-to-real results on Solo 8 robot. \textbf{Top}: Transition between walking and crawling; \textbf{Bottom}: Transition between walking and bipedalizing.}
\label{fig1}
 \vspace{-0.1cm}
\end{center}
    }]
}

\begin{document}
\maketitle
\thispagestyle{empty}
\pagestyle{empty}

%%%%%%%%%%%%%%%%%%%%%%%%%%%%%%%%%%%%%%%%%%%%%%%%%%%%%%%%%%%%%%%%%%%%%%%%%%%%%%%%
\begin{abstract}

Learning diverse skills for quadruped robots presents significant challenges, such as mastering complex transitions between different skills and handling tasks of varying difficulty. Existing imitation learning methods, while successful, rely on expensive datasets to reproduce expert behaviors. Inspired by introspective learning, we propose Progressive Adversarial Self-Imitation Skill Transition (PASIST), a novel method that eliminates the need for complete expert datasets. PASIST autonomously explores and selects high-quality trajectories based on predefined target poses instead of demonstrations, leveraging the Generative Adversarial Self-Imitation Learning (GASIL) framework. To further enhance learning, We develop a skill selection module to mitigate mode collapse by balancing the weights of skills with varying levels of difficulty. Through these methods, PASIST is able to reproduce skills corresponding to the target pose while achieving smooth and natural transitions between them. Evaluations on both simulation platforms and the Solo 8 robot confirm the effectiveness of PASIST, offering an efficient alternative to expert-driven learning.

\end{abstract}

%%%%%%%%%%%%%%%%%%%%%%%%%%%%%%%%%%%%%%%%%%%%%%%%%%%%%%%%%%%%%%%%%%%%%%%%%%%%%%%%
\section{INTRODUCTION}
Reinforcement Learning (RL) has demonstrated its effectiveness in learning diverse skills across domains such as robotics~\cite{10.1126/scirobotics.abk2822,zhuang2024humanoidparkourlearning}, autonomous driving~\cite{koirala2024f1tenthautonomousracingoffline, 10.1007/978-981-99-9119-8_22}, and financial risk management~\cite{fang2021universaltradingorderexecution}. In legged robots, especially with parallel simulation environments, RL enables robots to master complex skills in varied conditions in under half an hour~\cite{rudin2022learningwalkminutesusing}. However, traditional RL methods heavily depend on manually crafted reward functions specific to each task, increasing policy tuning complexity and requiring expert adjustment of reward weights~\cite{li2021reinforcement}.

To address these challenges, researchers have explored combining Imitation Learning (IL) with RL. IL enables robots to replicate expert behaviors using predefined reference datasets. Generative Adversarial Imitation Learning (GAIL, \cite{ho2016generative}) has inspired works focused on replacing complex reward functions with imitation-based approaches. These efforts have shown promise, including WASABI~\cite{li2023learning} and Cassi~\cite{10160421}. However, collecting expert datasets remains a major obstacle. Reference trajectories are typically captured using expensive motion capture systems from real animals~\cite{escontrela2022adversarialmotionpriorsmake} or generated through trajectory optimization~\cite{10160562}, both requiring extensive preprocessing. The former needs alignment for animal-robot morphological differences, while the latter is data-intensive. Moreover, tasks involving skill transitions, such as moving from quadrupedal to bipedal walking, often lack datasets for the transitions themselves, limiting efficient training when only skill-specific datasets are available.

Humans learn skills and transitions by focusing on target poses, imitating them through trial and error~\cite{huang2011rethinking}, and refining their movements for smooth transitions~\cite{feulner2025neural}, as in learning a complex dance sequence~\cite{serifi2024vmp}. We refer to this as \textbf{introspective learning}. This concept can be extended to robots by selecting high-quality exploration trajectories as imitation data, guiding skill learning toward more optimal behaviors. While existing IL methods rely on expert datasets, our approach, inspired by human introspective learning, limits imitation to physically realistic target poses and allows robots to autonomously generate motion features, enabling more natural transitions, which forms the basis of Generative Adversarial Self-Imitation Learning (GASIL)~\cite{guo2018generative}. However, GASIL can suffer from mode collapse, where policies overfit to sub-skills yielding high rewards. To address this, we introduce a skill selection module that calculates the skill selection command for the next timestep based on the performance of previous trajectories, balancing skill learning with varying difficulties.

We present the Progressive Adversarial Self-Imitation Skill Transition (PASIST) framework, inspired by human-like introspective learning for acquiring new skills. As shown in Fig.~\ref{fig1}, given only target poses, robots autonomously learn both the skills and their transitions by leveraging high-quality past trajectories. We also propose a new metric for evaluating high-quality trajectories, which combines task rewards and Dynamic Time Warping (DTW)~\cite{10.5555/3000850.3000887} values with target poses, incorporating human expert expectations. To our knowledge, this is the first application of the GASIL mechanism to skill learning in quadruped robots.

The main contributions of this work are as follows:
\begin{itemize}
    \item We develop a skill learning and transition framework based on GASIL, selecting high-quality past trajectories using task rewards and DTW values with target poses.
    \item We introduce a skill selector that balances skill learning by selecting the skills that require further training based on the performance of existing trajectories.
    \item We validate the effectiveness of PASIST through evaluations on simulation platforms and the Solo 8 robot.
\end{itemize}

\section{RELATED WORK}

\subsection{IL Applied to Robots}

IL has recently achieved significant breakthroughs in the area of quadruped robot locomotion. GAIL~\cite{ho2016generative} provides an effective approach for IL in high-dimensional environments by learning from state-action pairs in expert data, without relying on explicit cost functions. AMP~\cite{Peng_2021} extends this by utilizing expert data and task environments (even when only state trajectory data is available) to guide robots in task completion. AMP-based methods have been further applied to quadrupedal locomotion on complex terrains, demonstrating robust and agile capabilities~\cite{escontrela2022adversarialmotionpriorsmake,10167753}. Furthermore,~\cite{vollenweider2022advancedskillsmultipleadversarial} introduces Multi-AMP framework for developing multi-skill policies, integrating each expert dataset with corresponding generative adversarial networks to achieve more complex skill transfer and application. Cassi~\cite{10160421} combines Generative Adversarial Network (GAN) with unsupervised skill discovery techniques, enabling policies to imitate skills from expert data and addressing the issue of mode collapse that GAN might encounter when handling large unlabeled datasets and diverse skills. 

However, the quality of IL-based policies, such as AMP~\cite{Peng_2021}, heavily depends on the quality and completeness of the dataset. These datasets, which typically require extensive and precise construction, are often based on real animal motion segments or algorithmically generated data in simulation environments—both of which are time-consuming, resource-intensive, and difficult to scale efficiently to diverse tasks. In addition, IL-based methods cannot effectively extend actions beyond datasets.

\subsection{Self-Imitation Learning (SIL)}

Formally introduced in 2018, SIL~\cite{oh2018selfimitationlearning} provides a more flexible method for enhancing agent exploration by allowing agents to reinforce successful past behaviors. It has demonstrated exceptional performance across various RL environments, making it widely applicable to skill learning~\cite{NEURIPS2023_94796017}, large language model training~\cite{gulcehre2023reinforcedselftrainingrestlanguage}, and other downstream tasks~\cite{WANG2024111334}. SIL tackles the fundamental RL challenge of balancing exploration and exploitation by reproducing past successful experiences, which can indirectly lead to more effective exploration of the environment. The simplicity and adaptability of the SIL algorithm make it particularly suitable for integration into various actor-critic architectures, allowing for broad applicability~\cite{gangwani2019learningselfimitatingdiversepolicies}. When combined with GAN, SIL can be viewed as a generative adversarial extension of self-imitation learning. GASIL~\cite{guo2018generative} trains a discriminator to distinguish between the robot's current trajectory and past high-quality trajectories while simultaneously training the policy to make the discriminator unable to distinguish between the two. In scenarios with complex reward signals, GASIL helps address long-term credit assignment issues. GASIL, in particular, has been shown to be effective in diverse applications~\cite{xu2024harnessingnetworkeffectfake, zha2021douzeromasteringdoudizhuselfplay}. In multi-agent RL, GASIL is widely adopted as a baseline method because it facilitates enhanced exploration by integrating other exploration-boosting mechanisms like curiosity-driven approaches and hierarchical networks. This enables agents to learn from each other's successful trajectories, thereby improving collaboration and task completion in cooperative environments~\cite{GAILPG, hao2019independentgenerativeadversarialselfimitation}.

Despite successes of GASIL in other domains, its integration into quadruped robot training remains unexplored, where it could significantly enhance performance. Quadruped multi-skills learning involves extensive state and action spaces, with reward functions often being sparse, diverse, and non-generalizable, complicating training. PASIST addresses these challenges by reducing reliance on pre-constructed datasets, instead learning from autonomously discovered successful trajectories. This approach enables the robot to develop more natural, realistic behaviors and facilitates smoother transitions from simulation to real-world environments, a critical challenge in robotics.

\begin{figure*}[htbp]
\vspace{0.2cm}
\centerline{\includegraphics[width=\textwidth]{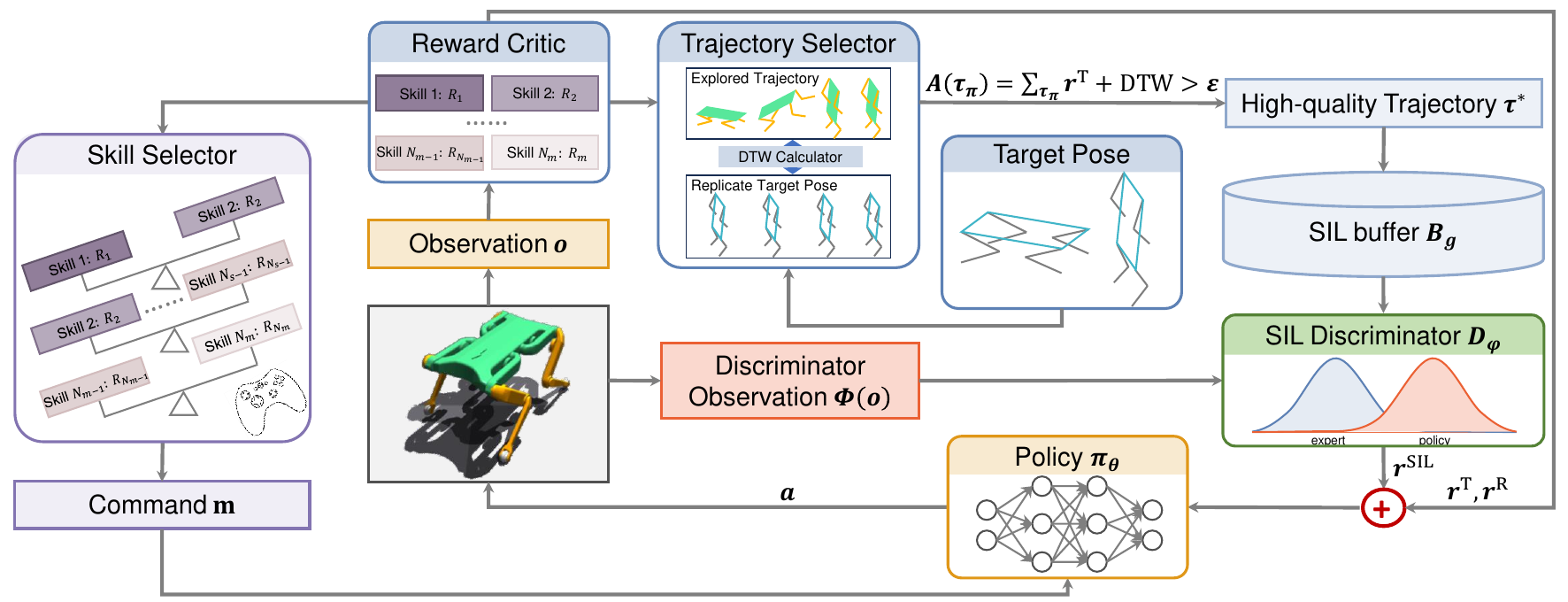}}
\captionsetup{font=small}
\caption{System overview. PASIST selects high-quality trajectories from policy exploration using the Trajectory Selector and performs imitation learning with the SIL discriminator. The SIL reward, combined with task and regularization rewards, is used to train the policy. A skill selector prevents overtraining on easier sub-policies, facilitating natural skill transitions without the need for expert datasets.}
\vspace{-0.6cm}
\label{fig2}
\end{figure*}

\section{METHOD}
In this section, we describe our method, PASIST, which autonomously explores to find high-quality trajectories for SIL without requiring guidance from an expert dataset. PASIST is inspired by GASIL~\cite{guo2018generativeadversarialselfimitationlearning}, and incorporates the work of AMP~\cite{Peng_2021}, which has been widely used for learning multi-skill movements from expert datasets. An overview of the method is illustrated in Figure~\ref{fig2}. In the following sections, we will briefly review the components of the approach.

\subsection{Learning Skills by Imitating High-quality Trajectories} 

The central concept of PASIST is that the robot should imitate the high-quality trajectories it has gathered. More specifically, following the GASIL framework, PASIST performs the following two updates for each iteration.

\subsubsection{Updating SIL buffer} 
PASIST maintains a trajectory buffer $B_g = \{\tau_i\}$ to collect high-quality trajectories $\tau_i^*$ for each skill $i$, which are represented as $\mathbf{\textup{o}}_t^I = \left[(o_{t-1}^{I}, o_{t}^{I}) |_{t=(1,2,3,\ldots)} \right] \in \mathcal{O}^I$, where $\mathcal{O}^I$ is the imitation observation space, consisting of the joint positions of the robot. 

Determining which past trajectories are high-quality is crucial to the effectiveness of learning. Intuitively, trajectories that achieve higher task rewards $r^{\text{T}}$ are more likely to align with the desired actions. However, relying solely on finely tuned rewards may not always produce the most desirable trajectories, as the robot might get stuck in local optima, leading to strange behaviors while attempting to maximize rewards. Additionally, the varying lengths of trajectories generated during the trial-and-error process make it difficult to accurately assess the quality of transitions between states based solely on task rewards. Therefore, it is more appropriate to incorporate trajectory similarity as one of the evaluation criteria for identifying high-quality trajectories. 

To achieve this, we take advantage of DTW value with an $L_2$ norm to compare the policy trajectory and the target pose $p$, matching and calculating distances between trajectories in a time-sequential manner. The target poses for different skills are collected from a trajectory optimization-based dataset~\cite{10160562} and stored in the target pose set $\mathcal{P}$, each skill corresponding to \textit{only one} target pose. They can also be obtained from real motion capture datasets or manually computed based on physical parameters.

After getting the target poses and high-quality trajectories, we compute the expected value, $\mathbb{E}[d^{\text{DTW}}(\Phi(\tau_{\pi}), \tau_{p})]$, as an evaluation metric, where $\tau_{\pi} \sim d_{\pi}$ represents a trajectory sampled from the policy's distribution, and $\Phi(\tau_{\pi})$ extracts the joint position information in $\mathcal{O}^I$ from the trajectory, which is convenient for DTW values calculation with $\tau_{p}$. Meanwhile, $\tau_{p}$ denotes the robot's reference trajectory, where the target pose is continuously duplicated along the time axis to match half the length of $\tau_{\pi}$.

Ultimately, combining task characteristics and trajectory similarity, we compute the assessment value:
\begin{equation}
\label{judgement}
    A(\tau_\pi) = \sum_{\tau_\pi}r^{\text{T}} + \mathbb{E}[d^{\text{DTW}}(\Phi(\tau_{\pi}), \tau_{p})],
\end{equation}
and use it as the criterion for evaluating high-quality trajectories. At the end of an episode, this value is calculated, and if it exceeds the maximum value $\epsilon$ in the SIL buffer, the trajectory is added to the buffer; otherwise, it is discarded. Note that to prevent catastrophic forgetting, the SIL buffer consistently maintains trajectories for all skill types, ensuring that the system retains knowledge across different skill domains.

\subsubsection{Updating SIL discriminator and policy}
In our experimental setup, the SIL discriminator is defined as a neural network parameterized by $\phi$. The objective of the SIL discriminator $D_\phi$, similar to~\cite{Peng_2021}, is to predict whether a transition $g(\mathbf{\textup{s}}) = \left[(s_{t-1}, s_t) |_{t=(1,2,3,\ldots)} \right] \in \mathcal{S}$ originates from a sample in the reference motion distribution $d^{\mathcal{M}}$ or from the policy transition distribution $d^\pi$,
\begin{equation}
    \begin{aligned}
    \mathbb{E}_{d^{\mathcal{M}}} \left [ \big(D_\phi( \mathbf{\textup{o}}_t^I) - 1 \big)^2 \right ] 
    & + \mathbb{E}_{ d^{\pi}} \left [ \big( D_\phi \left( g \left(\mathbf{\textup{s}} \right) \right)+ 1 \big)^2 \right ] \\
    & + \omega^{\text{GP}} \mathbb{E}_{d^{\mathcal{M}}} \left [ \big \|\nabla_{\mathbf{\textup{o}}_t^I} D_\phi \left (\mathbf{\textup{o}}_t^I \right) \big \|_2^2 \right ].
    \label{eqn:imitation_discriminator_loss}
    \end{aligned}
\end{equation}

Equation (\ref{eqn:imitation_discriminator_loss}) assigns a score of $+1$ to state transitions from the SIL buffer $B_g$, and $-1$ from the policy. The final term represents a penalty with weight $\omega_{\text{GP}}$, applied to non-zero gradients of samples from the SIL buffer, to stabilize training. The reward function used for policy training is
\begin{equation}
\label{SIL}
    r^{\text{SIL}} = \max \left[ 0, 1-0.25 \left( D_\phi(g(\mathbf{\textup{s}})) - 1 \right)^2 \right],
\end{equation}
which provides a reward in the range of $[0, 1]$, encouraging the learning of more challenging actions that are similar to those in the SIL buffer $B_g$.

\subsection{Preventing Mode Collapse in PASIST}

In policy training under the GAN framework, mode collapse is a common issue, often manifesting as the policy favoring sub-policies that yield higher rewards for specific skills. SIL method exacerbates this problem, as the randomness in acquiring high-quality trajectories makes it highly susceptible to getting stuck in local optima~\cite{gui2021review}. To address this, we develop a skill selector that balances the learning of skills with varying difficulty during training through adaptive command switching, ensuring satisfactory performance across all skills. Additionally, the robot can learn transitional movements between different skills without requiring a complete trajectory dataset for imitation.

To adaptively adjust the training iterations for different skills, we design a command selector $\mathbf{C}$, which consists of speed commands and motion skill commands, expressed as $\mathbf{C} = (\mathbf{v}, \mathbf{m})$. Here, $\mathbf{v} \in [-0.5, 0.5]$ represents the speed command, where random sampling trains the robot to perform skills at various speeds, enhancing policy robustness. $\mathbf{m}$ denotes the motion skill command in the command buffer $B_{\mathbf{C}}$, encoded as a one-hot vector with a dimension equal to the number of skills $\mathcal{N}_{\mathbf{m}}$, enabling more accurate computation and evaluation of task rewards $r^{\text{T}}$. Each command corresponds to a specific target pose, clearly identifying different commands for reward calculation. This module ensures that reward evaluation better reflects task requirements, improving training efficiency. The reward ratio for each command is calculated as:
\begin{equation}
\label{eq4}
\bar{r}^{\text{T},\mathbf{m}}_{t} = \frac{1}{\mathcal{N}_{\mathbf{x}}}  \sum_{\mathbf{x} \in B_\mathbf{C}} \left({r_{t}^{\text{T},\mathbf{x}}}\cdot\mathbb{I} \left(\mathbf{x} = \mathbf{m} \right) \right),
\end{equation}
where $\bar{r}^{\text{T},\mathbf{m}}_{t}$ represents the average reward for command $\mathbf{m}$. As the task reward $r^{\text{T}}$ is manually designed, we can estimate the reward for the optimal state $r^{\text{T},\mathbf{m}*}$ in advance, and use the following metric to evaluate the quality of skill training:
\begin{equation}
\label{eq5}
\mathbf{p} \left(\mathbf{x} = \mathbf{m} \right) = \frac{\bar{r}^{\text{T},\mathbf{m}}_{t}}{r^{\text{T},\mathbf{m}*}}.
\end{equation}

This metric determines which skill to train next. Higher $\mathbf{p}$ values reduce the likelihood of sampling subsequent skill commands corresponding to $\mathbf{m}'$, thereby achieving adaptive multi-skill training. If $\mathbf{m}' \neq \mathbf{m}$, this module, in combination with the GASIL framework, naturally guides and emphasizes the learning of transition policies between different skills.

\subsection{Rewards Formulation}

In PASIST, the reward function $r$ consists of three components: the SIL reward $r^{\text{SIL}}$, derived from the SIL discriminator for keyframe imitation, the task reward $r^\text{T}$, manually designed for different skills, and the regularization reward $r^{\text{R}}$. The total reward received by the policy is the sum of these three components:

\begin{equation}
\label{totalreward}
    r = \omega^{\text{SIL}} \omega^{\text{T}} r^{\text{SIL}} + (1 - \omega^{\text{T}}) r^{\text{T}} + \omega^{\text{R}} r^{\text{R}},
\end{equation}
where $\omega$ represents the weight of each reward component, with $\omega^{\text{R}} = 1.0$ remaining constant throughout the entire training process.
$\omega^{\text{SIL}}$ and $\omega^{\text{T}}$ are dynamically adjusted based on the task reward $r^{\text{T}}$ of each trajectory and its DTW value relative to the target pose. These adjustments determine whether the training objective in the next time period should prioritize imitating the target pose or exploring new actions, thereby addressing the fundamental RL challenge of balancing exploration and exploitation. The formal definitions of $\omega^{\text{SIL}}$ and $\omega^{\text{T}}$ are given as follows:
\begin{equation}
\label{parameter1}
    \omega^{\text{SIL}} = \frac{1}{\mathcal{N}_{\mathbf{m}}} \exp \left( - \sum_{p \in \mathcal{P}}
    \left\|\mathbb{E}[d^{\text{DTW}}(\Phi(\tau_{\text{SIL}}), \tau_{p})] - \sigma^{\text{SIL}} \right\| \right),
\end{equation}
\begin{equation}
\label{parameter2}
    \omega^{\text{T}} = \exp \left( - \left\| r^{\text{T}} - \sigma^{\text{T}}\right\| \right),
\end{equation}
where $\tau_{\text{SIL}}$ represents all high-quality trajectories stored in the SIL buffer, while $\sigma^{\text{SIL}}$ and $\sigma^{\text{T}}$ are two hyperparameters used to scale $\sigma^{\text{SIL}}$ and $\sigma^{\text{T}}$ within the range of $[0, 1]$. By dynamically adjusting the reward weights, the robot can autonomously balance imitation and exploration, achieving optimal performance within fewer timesteps.

Finally, we summarize PASIST in Algorithm~\ref{alg1}.
\begin{algorithm}[htb] 
\caption{PASIST} 
\footnotesize
\label{alg1} 
\begin{algorithmic}[1]
    \STATE \textbf{INPUT}: target poses set $\mathcal{P}$; feature map $\Phi$.
    \STATE \textbf{INITIALIZE}: policy parameter $\theta$; SIL discriminator parameter $\phi$; replay buffer $B \leftarrow \emptyset$; SIL buffer $B_g \leftarrow \emptyset$; maximum judgement value $\epsilon \leftarrow 0$.
    \FOR {learning iteration $ = 1,2,\dots$}
        \STATE Sample skill command $\mathbf{C}$ according to Eq.~\ref{eq4} and Eq.~\ref{eq5}.
        \STATE Collect trajectory $\tau_\pi=\left(s_0, a_0, r^T_0, s_1, a_1, r^T_1, \dots \right)$ with policy $\pi$.
        \STATE Extract the target pose $p \in \mathcal{P}$ corresponding to $\mathbf{C}$, and expand it into a trajectory $\tau_p$ with the half length as $\Phi(\tau_\pi)$.
        \STATE Calculate $A(\tau_\pi)$ and $r^{\text{SIL}}$ according to Equations~\ref{judgement} and~\ref{SIL}.
        \STATE Calculate total reward $r$ according to Equations \ref{totalreward}, \ref{parameter1}, and \ref{parameter2}.
        \STATE Fill replay buffer $B$ with $\tau_\pi$.
        \IF{$A(\tau_\pi)>\epsilon$}
            \STATE Fill SIL buffer with $\Phi(\tau_\pi)$.
            \STATE $\epsilon \leftarrow A(\tau_\pi)$.
        \ENDIF
        \FOR{policy learning epoch $ = 1,2,\dots$}
            \STATE Sample transition mini-batches $b^\pi\sim B$.
            \STATE Update $V$ and $\pi_\theta$ with PPO~\cite{PPO} or another RL algorithm.
        \ENDFOR
        \FOR{SIL discriminator learning epoch $ = 1,2,\dots$} 
            \STATE Sample transition mini-batches $b^\pi\sim B$ and $b^\mathcal{M} \sim B_g$.
            \STATE Update $d_\phi$ using $b^\pi$ and $b^\mathcal{M}$ according to Eq.~\ref{eqn:imitation_discriminator_loss}.
        \ENDFOR
    \ENDFOR 
\end{algorithmic}
\end{algorithm}
\vspace{-0.5cm}

\section{EXPERIMENTS}

\begin{figure*}[htbp]
\vspace{0.2cm}
\centerline{\includegraphics[width=\textwidth]{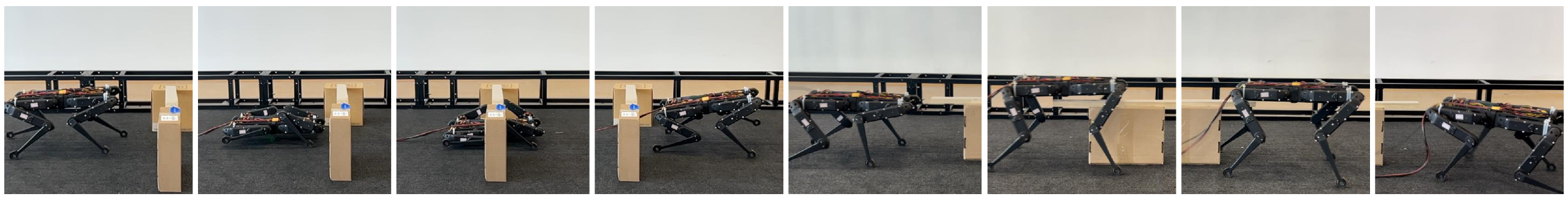}}
\captionsetup{font=small}
\caption{Sequences of Solo 8 robot navigating real-world obstacles by flexibly switching between learned skills.}
\vspace{-0.6cm}
\label{fig3}
\end{figure*}

The following experiments aim to address:
\begin{itemize}
    \item Can PASIST effectively train across diverse skills, each yielding favorable results? (In Section~\ref{problem1})
    \item Do the modules within PASIST contribute positively to the overall skill training process? (In Section~\ref{problem2})
    \item Can policies trained via PASIST successfully sim-to-real in a zero-shot manner? (In Section~\ref{problem3})
\end{itemize}

\subsection{Experimental Settings}

We use Isaac Gym~\cite{makoviychuk2021isaacgymhighperformance, liang2018gpuacceleratedroboticsimulationdistributed} as simulation platform, implementing PASIST with legged\_gym~\cite{rudin2022learningwalkminutesusing} and rsl\_rl~\cite{RSL-RL}. Both simulation and real-world testing are performed on the Solo 8 robot~\cite{Grimminger_2020}. Four skills and their transitions are implemented: \textbf{walk} ($\text{base\_height}=0.25$), \textbf{crawl} ($\text{base\_height}=0.1$), \textbf{stilt} ($\text{base\_height}=0.3$), and \textbf{bipedalize}. Each scenario is trained over 1,000 timesteps, focusing on two skills and their transitions. Due to the symmetric design of the Solo 8 robot, a single policy allows the robot to move both forward and backward while executing skills and transitions.

To facilitate policy transfer from simulation to real-world settings, domain randomization is employed. Table~\ref{table1} lists the randomized variables and their uniformly sampled ranges.

\begin{table}[h]
\captionsetup{font=small}
\vspace{-0.2cm}
\caption{Randomized Simulation Parameters.}
\vspace{-0.2cm}
\label{table1}
\begin{center}
\begin{tabular}{c|c}
\hline
\textbf{Parameter} & \textbf{Randomization Range}\\
\hline
Friction & $\left[0.2, 2.0 \right]$\\
Base Mass & $\left[-0.5, 0.5 \right]$kg\\
Position of CoM & $\left[-0.03, 0.03 \right]$m \\
Ground Height & $\left[-0.02, 0.02 \right]$m\\
Motor Gain Multiplier & $\left[0.8, 1.2 \right]$\\
Push Robots & Every 5s at $0.1$m/s\\
\hline
\end{tabular}
\vspace{-0.5cm}
\end{center}
\end{table}

\subsection{Imitation of Diverse Skills}
\label{problem1}
\subsubsection{Skills Distribution}
\label{DTW_similarity}

\begin{figure}[tbp]
\begin{minipage}[b]{0.48\linewidth}
  \centering
  \centerline{\includegraphics[width=4.2cm]{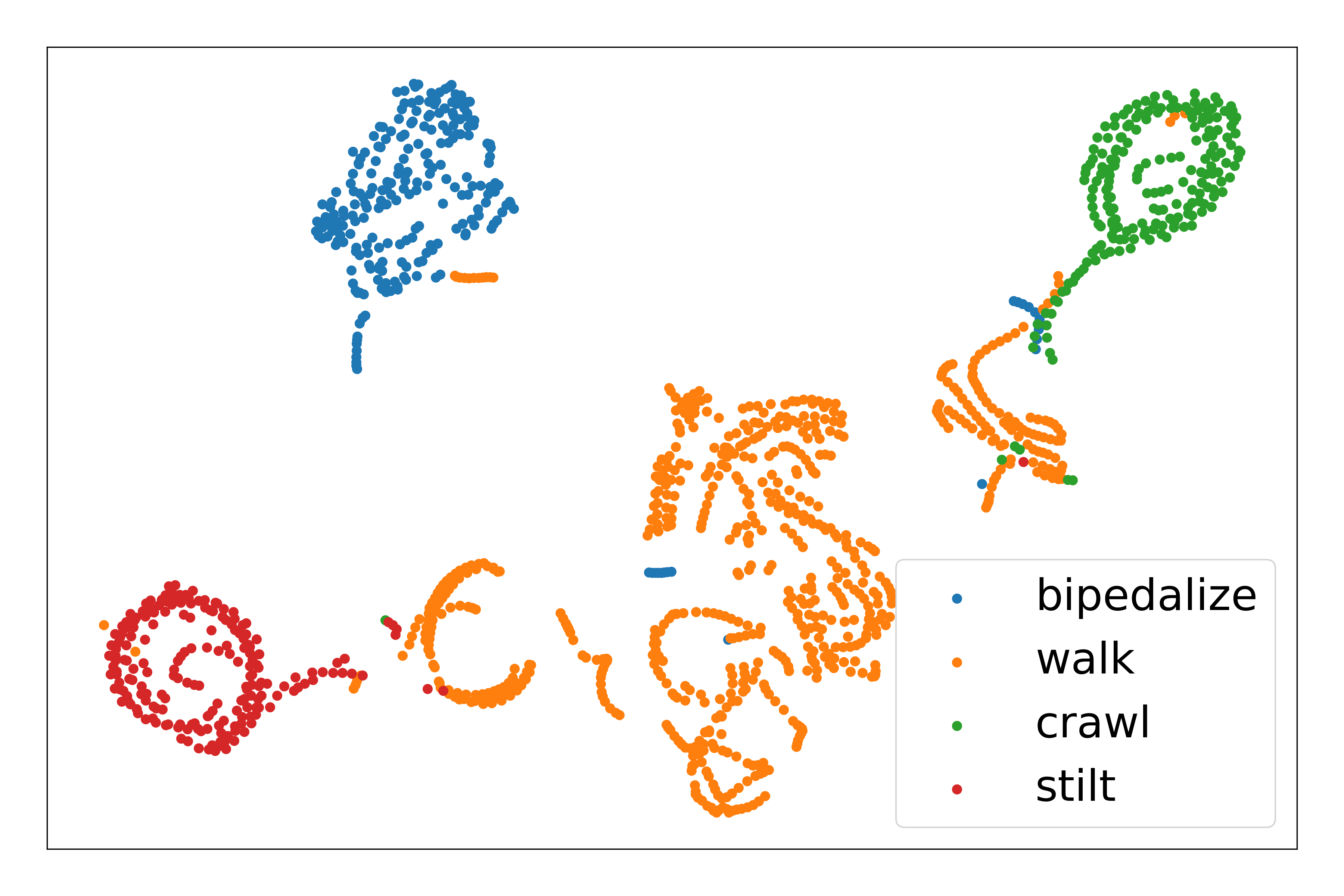}}
  \label{state-tsne}
\end{minipage}
\hfill
\begin{minipage}[b]{0.48\linewidth}
  \centering
  \centerline{\includegraphics[width=4.2cm]{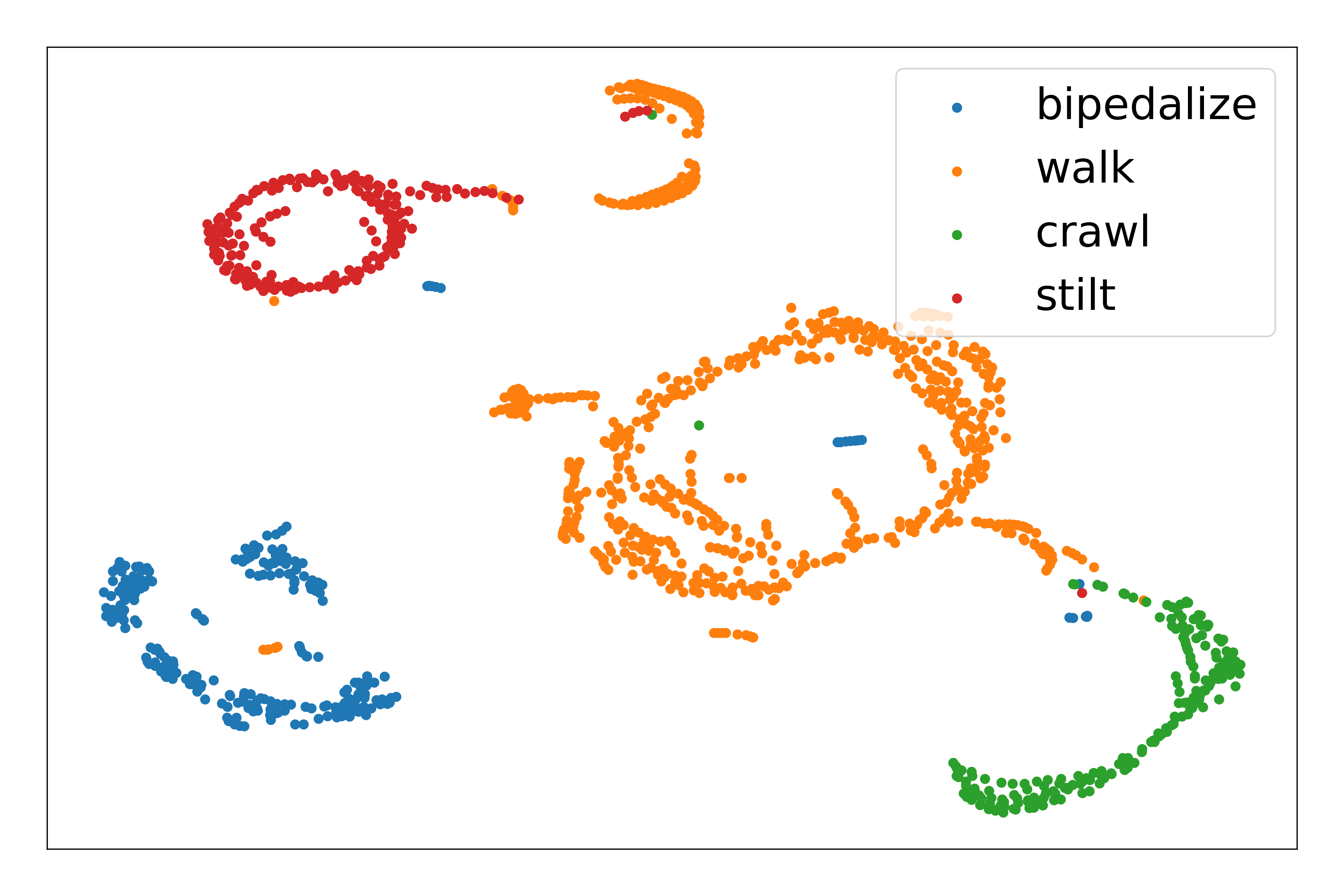}}
  \label{action-tsne}
\end{minipage}
\captionsetup{font=small}
\caption{t-SNE results of the state space (left) and action space (right) extracted from 500 steps of state and action data.}
\vspace{-0.5cm}
\label{tsne}
\end{figure}

This set of experiments aims to highlight the differences among the four skills. Four environment groups are established, each corresponding to a different locomotion skill: walk, crawl, stilt, and bipedalize. We collect 500 steps of state and action data generated by the joint positions. Intuitively, bipedalizing differs markedly from the other three skills. Due to the small contact area of the robot's feet, maintaining balance in the standing position requires rapid, small steps of the hind legs and continuous swinging of the front legs. In contrast, quadrupedal gaits, owing to their inherent stability, are more naturally learned. However, variations in base height, affecting stride length and balance, result in different walking dynamics. Fig.~\ref{tsne} displays the t-SNE visualization of the state and action spaces, revealing this phenomenon.

Table~\ref{table:dtw} presents the DTW values for dfferent skills using target pose, showing that DTW values among quadrupedal skills (walk, crawl, and stilt) are significantly lower compared to those between bipedalize and quadrupedal skills, indicating a higher similarity among quadrupedal skills. Overall, there are notable differences across all four skills.

\begin{table}[h]
\captionsetup{font=small}
\caption{DTW values between different skills.}
\vspace{-0.25cm}
\label{table:dtw}
\begin{center}
\begin{tabular}{>{\centering\arraybackslash}p{1.2cm}|>{\centering\arraybackslash}p{1.2cm}>{\centering\arraybackslash}p{1.2cm}>{\centering\arraybackslash}p{1.2cm}>{\centering\arraybackslash}p{1.2cm}}
\hline
& \textbf{walk} & \textbf{crawl} & \textbf{stilt} & \textbf{bipedalize} \\
\hline
\textbf{walk}  & 0  & 13.18 & 13.52 & 38.22 \\
\textbf{crawl} & 13.18  & 0 & 18.79 & 32.87 \\
\textbf{stilt} & 13.52 & 18.79 & 0 &  31.38 \\
\textbf{bipedalize} & 38.22 & 32.87 & 31.38 & 0 \\
\hline
\end{tabular}
\vspace{-0.25cm}
\end{center}
\end{table}

\subsubsection{Skills Imitation}

\begin{figure}[htbp]
\centerline{\includegraphics[width=\columnwidth]{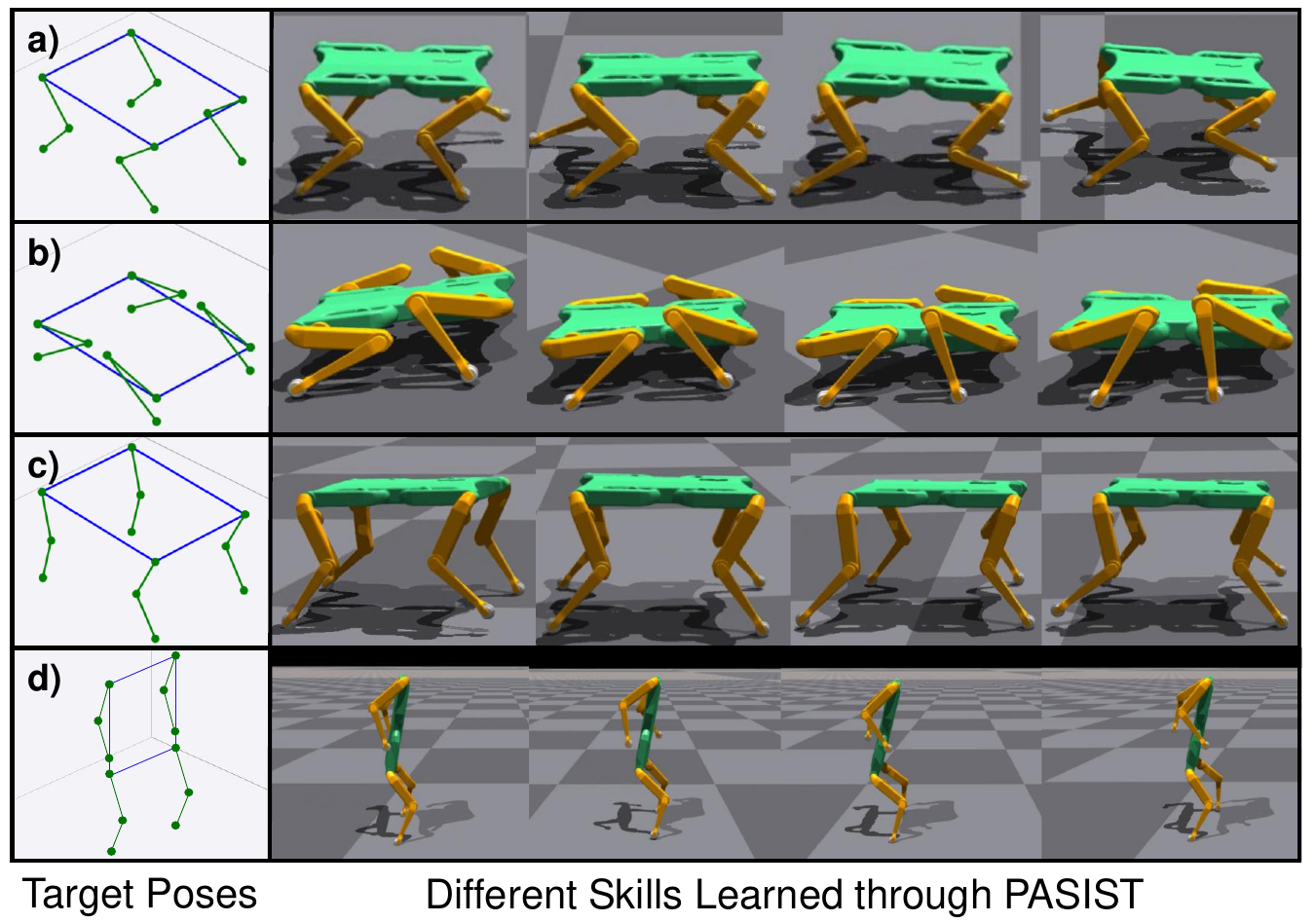}}
\captionsetup{font=small}
\caption{Comparison of the four skills learned through PASIST with their respective target poses. Subfigures a), b), c), and d) correspond to walk, crawl, stilt, and bipedalize, respectively.}
\vspace{-0.4cm}
\label{performance}
\end{figure}

The objective of this experiment is to demonstrate PASIST's ability to learn diverse skills. PASIST aims to enable the robot to acquire continuous actions based on target poses. To evaluate the performance of the algorithm, we present the four skills trained in the simulation environment and compare them with their respective target poses, as shown in Fig.~\ref{performance}. It can be observed that all four skills closely resemble their corresponding target poses, highlighting PASIST's effectiveness in multi-skill learning.

\subsection{Ablation Studies}
\label{problem2}
\begin{figure}[htbp]
\vspace{0.2cm}
\centerline{\includegraphics[width=7cm]{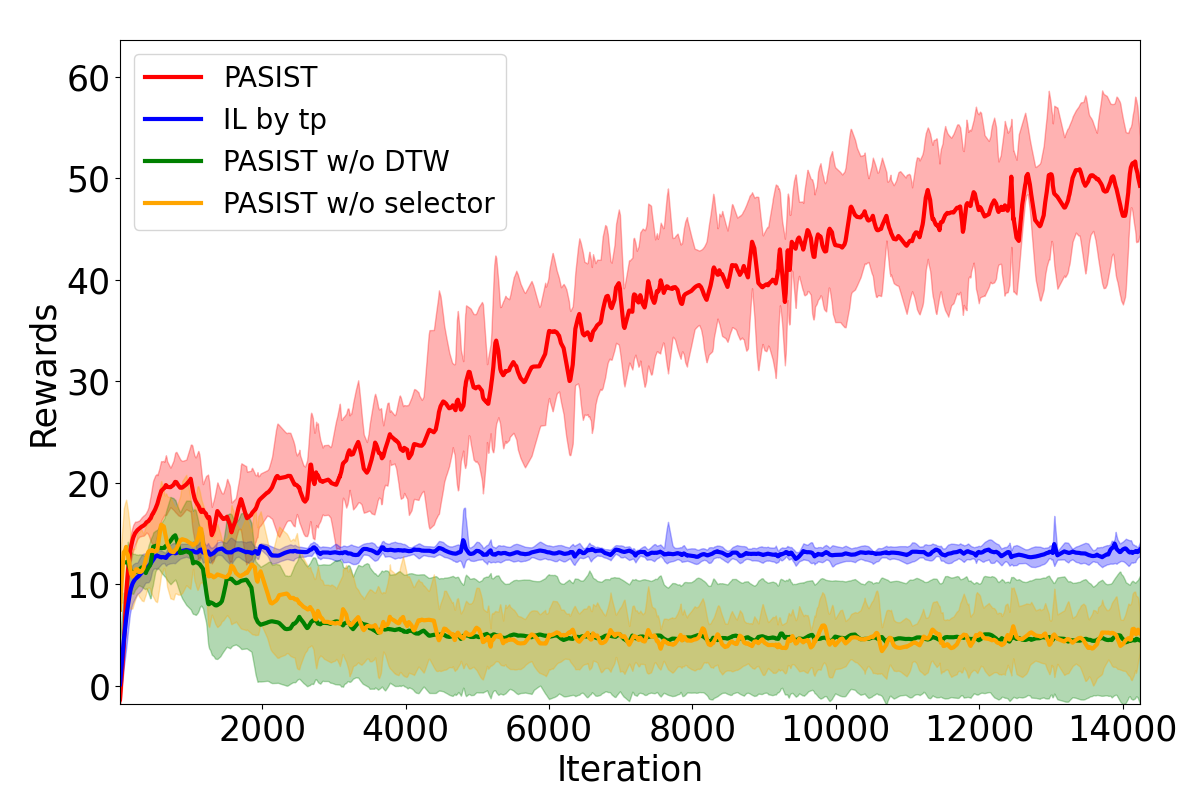}}
\captionsetup{font=small}
\caption{The variation curves for the total reward $r$ during training.}
\vspace{-0.6cm}
\label{reward_curve}
\end{figure}

To demonstrate the effectiveness of the different modules in PASIST, we conduct ablation studies on the skill transition between walking and bipedalizing, because this is the most challenging task in our experiment. The performance of PASIST is compared against the following three policies:

\begin{enumerate}
    \item \textbf{IL by tp}: Using the PASIST framework but trains using an imitation dataset created by copying the target pose (tp) to match half length of the sampled trajectory.
    \item \textbf{PASIST w/o DTW}: Using the PASIST framework but selects high-quality trajectories for the SIL buffer $B_g$ based solely on the highest task reward $r^\text{T}$, without applying DTW values for trajectory comparison.
    \item \textbf{PASIST w/o selector}: Using the PASIST framework but removes the skill selector, randomly choosing the skill to train at each timestep.
\end{enumerate}

Each policy is trained for 15,000 timesteps with 4 different random seeds, with all randomization settings kept identical during the training phase. The learning performance is evaluated based on the total reward $r$ accumulated throughout the training, shown in Eq.~\ref{totalreward}. During the testing phase, the robot's performance is assessed across 4,000 environments, where skill-switching commands are randomly introduced at different timesteps to test the robustness of the learned policies. The evaluation metric includes the similarity to the target pose, measured using DTW values.

\begin{figure}[htbp]
\centerline{\includegraphics[width=7cm]{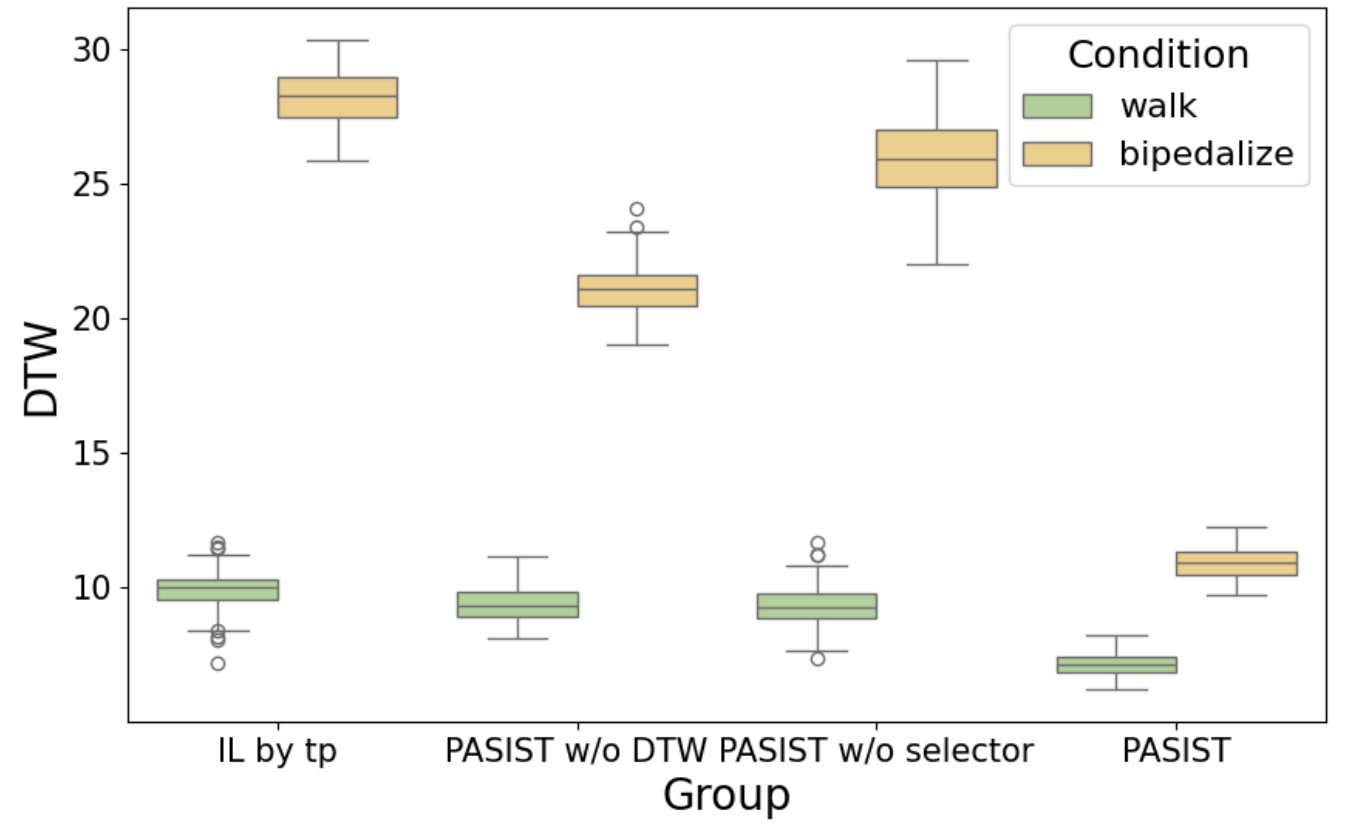}}
\captionsetup{font=small}
\caption{Two skills trained by different methods compared to the DTW values of their target poses. The lower the DTW value, the more similar the learned skill is to the target pose.}
\vspace{-0.2cm}
\label{DTW_box}
\end{figure}

The variation curves for the total reward $r$ are shown in Fig.~\ref{reward_curve}. PASIST consistently achieves the highest total reward, increasing over time and demonstrating its ability to continually learn from the SIL buffer. In contrast, other methods quickly converge to local optima within a few timesteps, hindering exploration. Additionally, as shown in Fig.~\ref{DTW_box}, all methods exhibit high similarity to the target pose in the walking skill. However, for the bipedalizing skill, only PASIST achieves satisfactory performance, while other methods produce poses that deviate significantly from the target. This highlights PASIST's superior capability in managing diverse skill transitions.

\begin{figure}[tbp]
\vspace{0.2cm}
\begin{minipage}[b]{0.48\linewidth}
  \centering
  \centerline{\includegraphics[width=4.2cm]{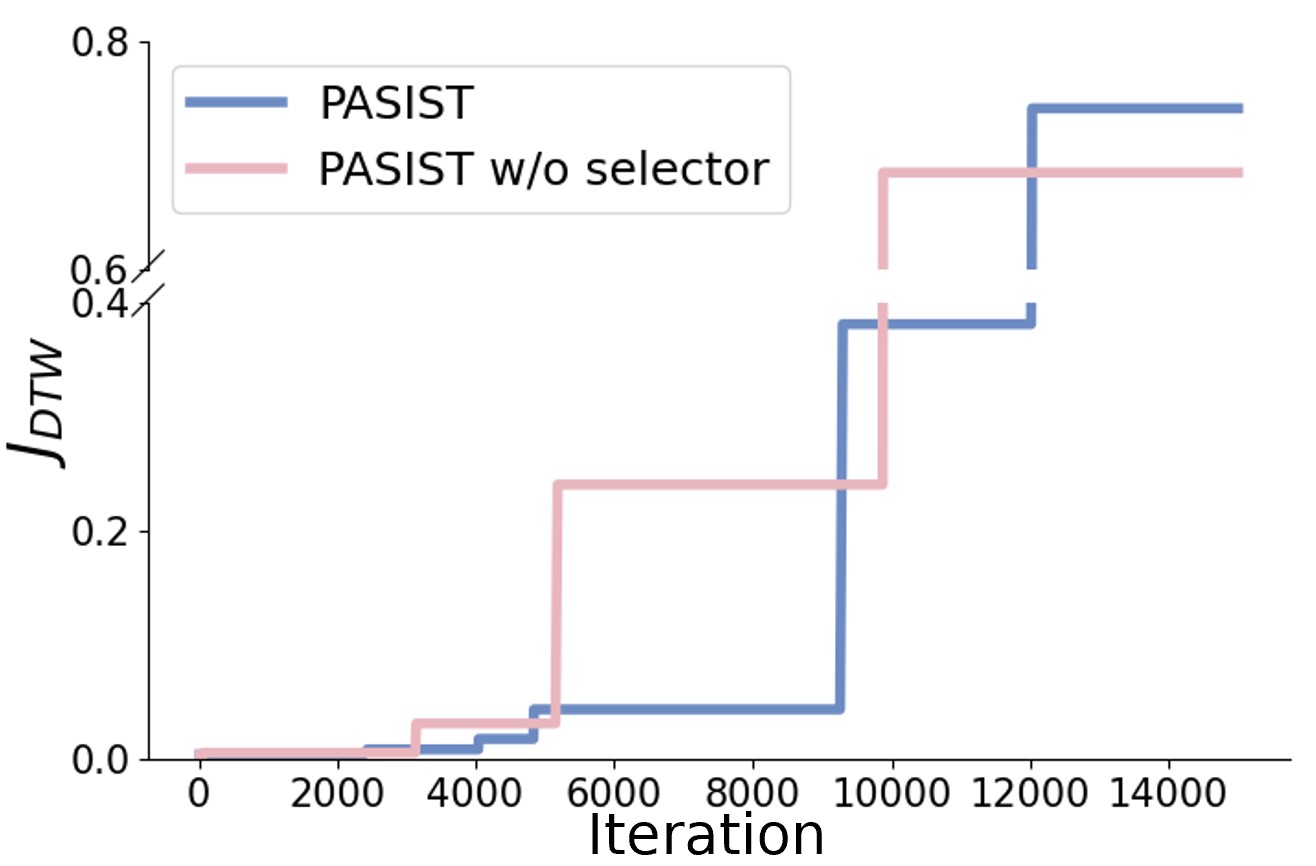}}
  \centerline{\small (a) walking.}\medskip
  \label{inference-time}
\end{minipage}
\hfill
\begin{minipage}[b]{0.48\linewidth}
  \centering
  \centerline{\includegraphics[width=4.2cm]{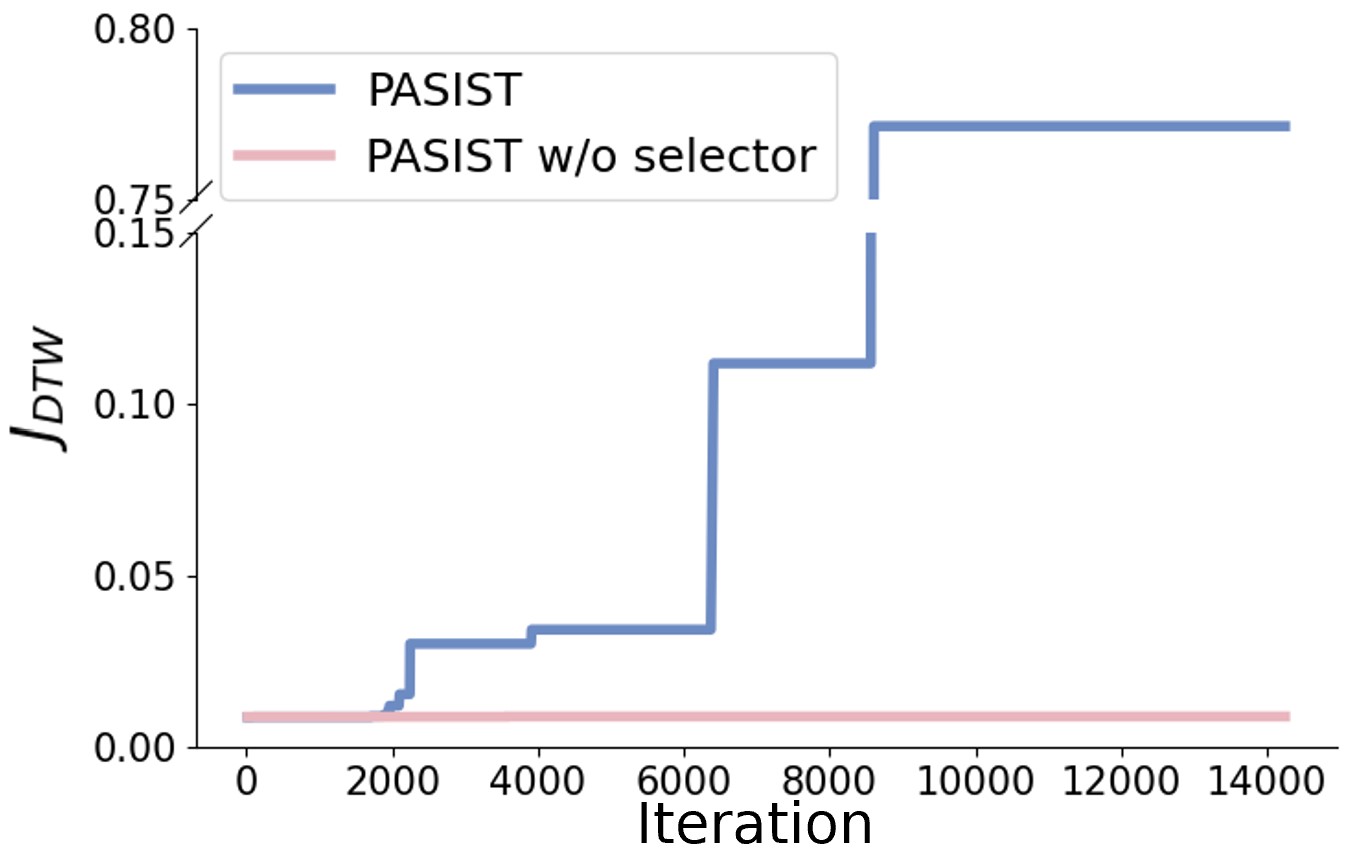}}
  \centerline{\small (b) bipedalizing.}\medskip
  \label{terrains}
\end{minipage}
\captionsetup{font=small}
\vspace{-0.3cm}
\caption{Variation of $J_\text{DTW}$ for high-quality trajectories of walking and bipedalizing skills stored in the SIL buffer during training. The higher the $J_\text{DTW}$, the higher the quality of the trajectory.}
\vspace{-0.6cm}
\label{J_DTW}
\end{figure}

Moreover, to further prove PASIST's effectiveness in mitigating mode collapse, we compare the quality of trajectories stored in the SIL buffer during training for both PASIST and PASIST w/o selector. We still focus on walking and bipedalizing, as they have the greatest morphological differences (Fig.~\ref{tsne}), and bipedalizing is significantly more challenging for quadruped robots. This makes it difficult for a single policy network to balance both skills. We evaluate these trajectories using a DTW-based metric, which measures their similarity to the target pose:
\begin{equation}
\label{DTW_judge}
    J_\text{DTW} = \exp \left( -\frac{\min \left( \left\| \mathbb{E}[d^{\text{DTW}}(\Phi(\tau_{\text{SIL}}), \tau_{p})] - \sigma^{\text{SIL}} \right\|, 0 \right)}{10}  \right),
\end{equation}
where $\sigma^{\text{SIL}}$ is the parameter. By Eq.~\ref{DTW_judge}, DTW values are normalized to the $[0, 1]$ range to adjust the granularity of SIL. A higher $J_\text{DTW}$ indicates a closer alignment with the target pose. Fig.~\ref{J_DTW} shows the results, where both methods capture high-quality trajectories for the walking skill, which is easier to learn. However, PASIST w/o selector struggles to extract effective trajectories for the bipedalizing skill, underscoring the importance of the skill selector module.

\subsection{Sim-to-real Transfer}
\label{problem3}

We deploy the policy on the real Solo 8 robot, as shown in Fig.~\ref{fig1}, achieving smooth transitions between walking and bipedalizing. Fig.~\ref{fig3} illustrates the robot executing various skills to navigate obstacles using a versatile policy. By switching policies, the robot effectively overcomes different obstacles. Compared with Fig.~\ref{performance}, the similar performance between simulation and real-world systems confirms the effectiveness of our zero-shot sim-to-real transfer.

\section{CONCLUSIONS}

In this work, we present PASIST, a novel approach enabling robots to explore and imitate high-quality trajectories based on target poses, eliminating the need for full expert datasets. PASIST combines task rewards with DTW values to assess trajectory quality and enhances training efficiency by using high-quality trajectories as the imitation dataset. It also addresses mode collapse by incorporating a skill selector that chooses the next skill to train based on the performance of learned policies. Our experiments show that PASIST effectively learns diverse skills and discovers smooth transitions, demonstrating strong generalization and robustness. Future work will explore the integration of additional sensor data and the application of PASIST to more complex tasks like navigation, further enhancing its real-world applicability.


%%%%%%%%%%%%%%%%%%%%%%%%%%%%%%%%%%%%%%%%%%%%%%%%%%%%%%%%%%%%%%%%%%%%%%%%%%%%%%%%

%%%%%%%%%%%%%%%%%%%%%%%%%%%%%%%%%%%%%%%%%%%%%%%%%%%%%%%%%%%%%%%%%%%%%%%%%%%%%%%%

%%%%%%%%%%%%%%%%%%%%%%%%%%%%%%%%%%%%%%%%%%%%%%%%%%%%%%%%%%%%%%%%%%%%%%%%%%%%%%%%

\bibliographystyle{IEEEtran}
\bibliography{mylib}

\begin{thebibliography}{10}
\providecommand{\url}[1]{#1}
\csname url@samestyle\endcsname
\providecommand{\newblock}{\relax}
\providecommand{\bibinfo}[2]{#2}
\providecommand{\BIBentrySTDinterwordspacing}{\spaceskip=0pt\relax}
\providecommand{\BIBentryALTinterwordstretchfactor}{4}
\providecommand{\BIBentryALTinterwordspacing}{\spaceskip=\fontdimen2\font plus
\BIBentryALTinterwordstretchfactor\fontdimen3\font minus \fontdimen4\font\relax}
\providecommand{\BIBforeignlanguage}[2]{{%
\expandafter\ifx\csname l@#1\endcsname\relax
\typeout{** WARNING: IEEEtran.bst: No hyphenation pattern has been}%
\typeout{** loaded for the language `#1'. Using the pattern for}%
\typeout{** the default language instead.}%
\else
\language=\csname l@#1\endcsname
\fi
#2}}
\providecommand{\BIBdecl}{\relax}
\BIBdecl

\bibitem{10.1126/scirobotics.abk2822}
\BIBentryALTinterwordspacing
T.~Miki, J.~Lee, J.~Hwangbo, L.~Wellhausen, V.~Koltun, and M.~Hutter, ``Learning robust perceptive locomotion for quadrupedal robots in the wild,'' \emph{Science Robotics}, vol.~7, no.~62, p. eabk2822, 2022. [Online]. Available: \url{https://www.science.org/doi/abs/10.1126/scirobotics.abk2822}
\BIBentrySTDinterwordspacing

\bibitem{zhuang2024humanoidparkourlearning}
\BIBentryALTinterwordspacing
Z.~Zhuang, S.~Yao, and H.~Zhao, ``Humanoid parkour learning,'' 2024. [Online]. Available: \url{https://arxiv.org/abs/2406.10759}
\BIBentrySTDinterwordspacing

\bibitem{koirala2024f1tenthautonomousracingoffline}
\BIBentryALTinterwordspacing
P.~Koirala and C.~Fleming, ``F1tenth autonomous racing with offline reinforcement learning methods,'' 2024. [Online]. Available: \url{https://arxiv.org/abs/2408.04198}
\BIBentrySTDinterwordspacing

\bibitem{10.1007/978-981-99-9119-8_22}
X.~Wei, T.~Hou, X.~Zhao, J.~Tu, H.~Guan, P.~Zhai, and L.~Zhang, ``Reinforcement learning-based algorithm for real-time automated parking decision making,'' in \emph{Artificial Intelligence}, L.~Fang, J.~Pei, G.~Zhai, and R.~Wang, Eds.\hskip 1em plus 0.5em minus 0.4em\relax Singapore: Springer Nature Singapore, 2024, pp. 242--252.

\bibitem{fang2021universaltradingorderexecution}
\BIBentryALTinterwordspacing
Y.~Fang, K.~Ren, W.~Liu, D.~Zhou, W.~Zhang, J.~Bian, Y.~Yu, and T.-Y. Liu, ``Universal trading for order execution with oracle policy distillation,'' 2021. [Online]. Available: \url{https://arxiv.org/abs/2103.10860}
\BIBentrySTDinterwordspacing

\bibitem{rudin2022learningwalkminutesusing}
\BIBentryALTinterwordspacing
N.~Rudin, D.~Hoeller, P.~Reist, and M.~Hutter, ``Learning to walk in minutes using massively parallel deep reinforcement learning,'' 2022. [Online]. Available: \url{https://arxiv.org/abs/2109.11978}
\BIBentrySTDinterwordspacing

\bibitem{li2021reinforcement}
Z.~Li, X.~Cheng, X.~B. Peng, P.~Abbeel, S.~Levine, G.~Berseth, and K.~Sreenath, ``Reinforcement learning for robust parameterized locomotion control of bipedal robots,'' in \emph{2021 IEEE International Conference on Robotics and Automation (ICRA)}.\hskip 1em plus 0.5em minus 0.4em\relax IEEE, 2021, pp. 2811--2817.

\bibitem{ho2016generative}
J.~Ho and S.~Ermon, ``Generative adversarial imitation learning,'' \emph{Advances in neural information processing systems}, vol.~29, 2016.

\bibitem{li2023learning}
C.~Li, M.~Vlastelica, S.~Blaes, J.~Frey, F.~Grimminger, and G.~Martius, ``Learning agile skills via adversarial imitation of rough partial demonstrations,'' in \emph{Conference on Robot Learning}.\hskip 1em plus 0.5em minus 0.4em\relax PMLR, 2023, pp. 342--352.

\bibitem{10160421}
C.~Li, S.~Blaes, P.~Kolev, M.~Vlastelica, J.~Frey, and G.~Martius, ``Versatile skill control via self-supervised adversarial imitation of unlabeled mixed motions,'' in \emph{2023 IEEE International Conference on Robotics and Automation (ICRA)}, 2023, pp. 2944--2950.

\bibitem{escontrela2022adversarialmotionpriorsmake}
\BIBentryALTinterwordspacing
A.~Escontrela, X.~B. Peng, W.~Yu, T.~Zhang, A.~Iscen, K.~Goldberg, and P.~Abbeel, ``Adversarial motion priors make good substitutes for complex reward functions,'' 2022. [Online]. Available: \url{https://arxiv.org/abs/2203.15103}
\BIBentrySTDinterwordspacing

\bibitem{10160562}
Y.~Fuchioka, Z.~Xie, and M.~Van~de Panne, ``Opt-mimic: Imitation of optimized trajectories for dynamic quadruped behaviors,'' in \emph{2023 IEEE International Conference on Robotics and Automation (ICRA)}, 2023, pp. 5092--5098.

\bibitem{huang2011rethinking}
V.~S. Huang, A.~Haith, P.~Mazzoni, and J.~W. Krakauer, ``Rethinking motor learning and savings in adaptation paradigms: model-free memory for successful actions combines with internal models,'' \emph{Neuron}, vol.~70, no.~4, pp. 787--801, 2011.

\bibitem{feulner2025neural}
B.~Feulner, M.~G. Perich, L.~E. Miller, C.~Clopath, and J.~A. Gallego, ``A neural implementation model of feedback-based motor learning,'' \emph{Nature Communications}, vol.~16, no.~1, p. 1805, 2025.

\bibitem{serifi2024vmp}
A.~Serifi, R.~Grandia, E.~Knoop, M.~Gross, and M.~B{\"a}cher, ``Vmp: Versatile motion priors for robustly tracking motion on physical characters,'' in \emph{Computer Graphics Forum}, vol.~43, no.~8.\hskip 1em plus 0.5em minus 0.4em\relax Wiley Online Library, 2024, p. e15175.

\bibitem{guo2018generative}
Y.~Guo, J.~Oh, S.~Singh, and H.~Lee, ``Generative adversarial self-imitation learning,'' \emph{arXiv preprint arXiv:1812.00950}, 2018.

\bibitem{10.5555/3000850.3000887}
D.~J. Berndt and J.~Clifford, ``Using dynamic time warping to find patterns in time series,'' in \emph{Proceedings of the 3rd International Conference on Knowledge Discovery and Data Mining}, ser. AAAIWS'94.\hskip 1em plus 0.5em minus 0.4em\relax AAAI Press, 1994, p. 359–370.

\bibitem{Peng_2021}
\BIBentryALTinterwordspacing
X.~B. Peng, Z.~Ma, P.~Abbeel, S.~Levine, and A.~Kanazawa, ``Amp: adversarial motion priors for stylized physics-based character control,'' \emph{ACM Transactions on Graphics}, vol.~40, no.~4, p. 1–20, Jul. 2021. [Online]. Available: \url{http://dx.doi.org/10.1145/3450626.3459670}
\BIBentrySTDinterwordspacing

\bibitem{10167753}
J.~Wu, G.~Xin, C.~Qi, and Y.~Xue, ``Learning robust and agile legged locomotion using adversarial motion priors,'' \emph{IEEE Robotics and Automation Letters}, vol.~8, no.~8, pp. 4975--4982, 2023.

\bibitem{vollenweider2022advancedskillsmultipleadversarial}
\BIBentryALTinterwordspacing
E.~Vollenweider, M.~Bjelonic, V.~Klemm, N.~Rudin, J.~Lee, and M.~Hutter, ``Advanced skills through multiple adversarial motion priors in reinforcement learning,'' 2022. [Online]. Available: \url{https://arxiv.org/abs/2203.14912}
\BIBentrySTDinterwordspacing

\bibitem{oh2018selfimitationlearning}
\BIBentryALTinterwordspacing
J.~Oh, Y.~Guo, S.~Singh, and H.~Lee, ``Self-imitation learning,'' 2018. [Online]. Available: \url{https://arxiv.org/abs/1806.05635}
\BIBentrySTDinterwordspacing

\bibitem{NEURIPS2023_94796017}
R.~Huang, X.~Wu, H.~Yu, Z.~Fan, H.~Fu, Q.~Fu, and W.~Yang, ``A robust and opponent-aware league training method for starcraft ii,'' in \emph{Advances in Neural Information Processing Systems}, A.~Oh, T.~Naumann, A.~Globerson, K.~Saenko, M.~Hardt, and S.~Levine, Eds.\hskip 1em plus 0.5em minus 0.4em\relax Curran Associates, Inc., 2023, pp. 47\,554--47\,574.

\bibitem{gulcehre2023reinforcedselftrainingrestlanguage}
\BIBentryALTinterwordspacing
C.~Gulcehre, T.~L. Paine, S.~Srinivasan, K.~Konyushkova, L.~Weerts, A.~Sharma, A.~Siddhant, A.~Ahern, M.~Wang, C.~Gu, W.~Macherey, A.~Doucet, O.~Firat, and N.~de~Freitas, ``Reinforced self-training (rest) for language modeling,'' 2023. [Online]. Available: \url{https://arxiv.org/abs/2308.08998}
\BIBentrySTDinterwordspacing

\bibitem{WANG2024111334}
\BIBentryALTinterwordspacing
G.~Wang, F.~Wu, X.~Zhang, N.~Guo, and Z.~Zheng, ``Adaptive trajectory-constrained exploration strategy for deep reinforcement learning,'' \emph{Knowledge-Based Systems}, vol. 285, p. 111334, 2024. [Online]. Available: \url{https://www.sciencedirect.com/science/article/pii/S0950705123010821}
\BIBentrySTDinterwordspacing

\bibitem{gangwani2019learningselfimitatingdiversepolicies}
\BIBentryALTinterwordspacing
T.~Gangwani, Q.~Liu, and J.~Peng, ``Learning self-imitating diverse policies,'' 2019. [Online]. Available: \url{https://arxiv.org/abs/1805.10309}
\BIBentrySTDinterwordspacing

\bibitem{xu2024harnessingnetworkeffectfake}
\BIBentryALTinterwordspacing
X.~Xu, K.~Deng, M.~Dann, and X.~Zhang, ``Harnessing network effect for fake news mitigation: Selecting debunkers via self-imitation learning,'' 2024. [Online]. Available: \url{https://arxiv.org/abs/2402.03357}
\BIBentrySTDinterwordspacing

\bibitem{zha2021douzeromasteringdoudizhuselfplay}
\BIBentryALTinterwordspacing
D.~Zha, J.~Xie, W.~Ma, S.~Zhang, X.~Lian, X.~Hu, and J.~Liu, ``Douzero: Mastering doudizhu with self-play deep reinforcement learning,'' 2021. [Online]. Available: \url{https://arxiv.org/abs/2106.06135}
\BIBentrySTDinterwordspacing

\bibitem{GAILPG}
W.~Li, S.~Huang, Z.~Qiu, and A.~Song, ``Gailpg: Multi-agent policy gradient with generative adversarial imitation learning,'' \emph{IEEE Transactions on Games}, vol.~PP, pp. 1--14, 01 2024.

\bibitem{hao2019independentgenerativeadversarialselfimitation}
\BIBentryALTinterwordspacing
X.~Hao, W.~Wang, J.~Hao, and Y.~Yang, ``Independent generative adversarial self-imitation learning in cooperative multiagent systems,'' 2019. [Online]. Available: \url{https://arxiv.org/abs/1909.11468}
\BIBentrySTDinterwordspacing

\bibitem{guo2018generativeadversarialselfimitationlearning}
\BIBentryALTinterwordspacing
Y.~Guo, J.~Oh, S.~Singh, and H.~Lee, ``Generative adversarial self-imitation learning,'' 2018. [Online]. Available: \url{https://arxiv.org/abs/1812.00950}
\BIBentrySTDinterwordspacing

\bibitem{gui2021review}
J.~Gui, Z.~Sun, Y.~Wen, D.~Tao, and J.~Ye, ``A review on generative adversarial networks: Algorithms, theory, and applications,'' \emph{IEEE transactions on knowledge and data engineering}, vol.~35, no.~4, pp. 3313--3332, 2021.

\bibitem{PPO}
\BIBentryALTinterwordspacing
J.~Schulman, F.~Wolski, P.~Dhariwal, A.~Radford, and O.~Klimov, ``Proximal policy optimization algorithms,'' \emph{CoRR}, vol. abs/1707.06347, 2017. [Online]. Available: \url{http://arxiv.org/abs/1707.06347}
\BIBentrySTDinterwordspacing

\bibitem{makoviychuk2021isaacgymhighperformance}
\BIBentryALTinterwordspacing
V.~Makoviychuk, L.~Wawrzyniak, Y.~Guo, M.~Lu, K.~Storey, M.~Macklin, D.~Hoeller, N.~Rudin, A.~Allshire, A.~Handa, and G.~State, ``Isaac gym: High performance gpu-based physics simulation for robot learning,'' 2021. [Online]. Available: \url{https://arxiv.org/abs/2108.10470}
\BIBentrySTDinterwordspacing

\bibitem{liang2018gpuacceleratedroboticsimulationdistributed}
\BIBentryALTinterwordspacing
J.~Liang, V.~Makoviychuk, A.~Handa, N.~Chentanez, M.~Macklin, and D.~Fox, ``Gpu-accelerated robotic simulation for distributed reinforcement learning,'' 2018. [Online]. Available: \url{https://arxiv.org/abs/1810.05762}
\BIBentrySTDinterwordspacing

\bibitem{RSL-RL}
\BIBentryALTinterwordspacing
E.~Robotic Systems~Lab, ``rsl\_rl,'' 2024. [Online]. Available: \url{\url{https://github.com/leggedrobotics/rsl\_rl}}
\BIBentrySTDinterwordspacing

\bibitem{Grimminger_2020}
\BIBentryALTinterwordspacing
F.~Grimminger, A.~Meduri, M.~Khadiv, J.~Viereck, M.~Wuthrich, M.~Naveau, V.~Berenz, S.~Heim, F.~Widmaier, T.~Flayols, J.~Fiene, A.~Badri-Sprowitz, and L.~Righetti, ``An open torque-controlled modular robot architecture for legged locomotion research,'' \emph{IEEE Robotics and Automation Letters}, vol.~5, no.~2, p. 3650–3657, Apr. 2020. [Online]. Available: \url{http://dx.doi.org/10.1109/LRA.2020.2976639}
\BIBentrySTDinterwordspacing

\end{thebibliography}

\end{document}